\documentclass[11pt]{article}
\usepackage[utf8]{inputenc}
\usepackage{amsmath, amssymb, amsfonts}
\usepackage{geometry}
\usepackage{graphicx}
\usepackage{hyperref}
\usepackage{natbib}
\usepackage{times}
\usepackage{booktabs}
\usepackage{fancyhdr}

\hypersetup{colorlinks=true, urlcolor=blue, citecolor=black}
\title{Adaptive Thresholding for Multi-Label Classification via Global-Local Signal Fusion}
\author{
  Dmytro Shamatrin\thanks{
    Work conducted independently; current affiliation: Oracle. 
    \href{https://orcid.org/0009-0003-9497-2395}{ORCID: 0009-0003-9497-2395}.\\
    Draft version 1.7, submitted to arXiv.
  } \\
  Independent Researcher \\
  \texttt{dmytro.shamatrin@gmail.com}
}

\date{05-May-2025}

\begin{document}

\maketitle

\begin{abstract}
Multi-label classification (MLC) requires predicting multiple labels per sample, often under heavy class imbalance and noisy conditions. 
Traditional approaches apply fixed thresholds or treat labels independently, overlooking context and global rarity. 
We introduce an adaptive thresholding mechanism that fuses global (IDF-based) and local (KNN-based) signals to produce per-label, per-instance thresholds. 
Instead of applying these as hard cutoffs, we treat them as differentiable penalties in the loss, providing smooth supervision and better calibration. 
Our architecture is lightweight, interpretable, and highly modular. 
On the AmazonCat-13K benchmark, it achieves a macro-F1 of 0.1712, substantially outperforming tree-based and pretrained transformer-based methods.
We release full code for reproducibility and future extensions.
\end{abstract}

\section{Introduction}
Multi-label classification (MLC) demands simultaneous prediction of multiple labels, often in imbalanced and noisy contexts.
In domains like medicine, labels vary dramatically in severity and diagnostic consequence.
Applying a single uniform threshold across all labels is not only naive but potentially harmful in high-stakes settings.
Our work proposes a novel adaptive thresholding function that fuses global and local information to dynamically penalize logits per label, per sample, rather than directly thresholding probabilities.

While our method was initially motivated by clinical tasks such as ICD code assignment, we first validated it on AmazonCat-13K. To support reproducibility and encourage adoption, we publicly released our implementation as the \texttt{MLC Adaptive Thresholding} toolkit~\citep{mlcadaptive}.

Adaptive thresholding provides a principled way to manage this complexity, but existing methods often treat each label or instance in isolation.
We argue that real-world tasks require thresholds that consider both global patterns of label scarcity and local evidence of similarity.
Our approach builds on this insight, and we aim to deliver a more stable, interpretable, and context-aware thresholding method tailored for extreme multi-label classification.

\section{Related Work}
Static thresholding methods often fail to capture contextual sensitivity, relying on global heuristics such as the commonly used 0.5 cutoff.
Adaptive thresholding has seen progress with label-specific optimization or learnable thresholds, yet instance-level thresholding remains relatively unexplored.

A complementary line of research proposes embedding output labels into continuous spaces to model label similarity directly~\citep{NIPS2014_94b80c78}. 
While our method retains a discrete label representation, it introduces label-aware threshold modulation via global and local signals, offering an interpretable and modular alternative to label embeddings.

KNN-based local learning heuristics offer a promising route for modeling instance context, but are seldom integrated into learnable systems. Though early notions of local thresholding have appeared in informal discussions, formal integration into end-to-end learnable systems remains rare.
Likewise, IDF-based representations have proven useful for measuring label rarity in text-based MLC, though rarely for modulating thresholds.
Attention-based architectures offer an alternative, but often sacrifice interpretability.
Recent advances in pseudo-label augmentation and long-tailed label calibration~\citep{zhang2022longtailedextrememultilabeltext} explore label-wise adaptivity, but do not explicitly model fusion between global label statistics and local context signals.

Tree-based models such as AttentionXML \citep{you2019attentionxmllabeltreebasedattentionaware} show strong performance in extreme MLC tasks but rely on label hierarchies, while datasets like RCV1 \citep{lewis2004rcv1} have been widely used but lack the label scale and imbalance of benchmarks like AmazonCat-13K.
Earlier work explored class-calibrated heuristics and threshold adjustment, but lacked end-to-end training or fusion of global and local context.
Foundational designs such as character-level CNNs \citep{zhang2016characterlevelconvolutionalnetworkstext} and adversarial autoencoders \citep{makhzani2016adversarialautoencoders} helped shape regularization and feature control strategies that influence our architecture.

While we do not use such architectures directly, our design benefits from their underlying principles of representation shaping and activation margin control.
Some earlier work explored deep CNNs for extreme MLC tasks \citep{10.1145/3077136.3080834}, but did not incorporate label-specific thresholds or fuse global and local cues, while earlier efforts like KAN and RinSCut \citep{lee2002kanrinscut} attempted local thresholding heuristics.

\section{Methodology}
Let $\text{IDF}_l$ denote a global rarity score of label $l$\footnote{This corresponds to the inverse document frequency component in traditional TF-IDF, computed globally over label occurrence.}, and $\text{KNN}_l(x)$ denote a local agreement score from neighboring instances.

We define the adaptive threshold $\theta_l(x)$ as:
\begin{equation}
\theta_l(x) = \lambda \cdot \alpha_l \cdot \text{IDF}_l + (1 - \lambda) \cdot \beta_l \cdot \text{KNN}_l(x) + b_l
\end{equation}
where $\lambda$ is a learnable blend weight, $\alpha_l, \beta_l$ are signal importance weights, and $b_l$ is a label-specific bias. Optionally, logits may be standardized before loss application:
\begin{equation}
\hat{z}_l = \frac{z_l - \mu}{\sigma + \epsilon}
\end{equation}

We apply a composite loss:
\begin{equation}
\mathcal{L}(x) = \sum_l \text{BCEWithLogits}(z_l(x) - \theta_l(x), y_l) + \lambda_m \cdot \text{MarginLoss}(z_l(x), \theta_l(x), y_l)
\end{equation}

\textbf{Local KNN Signal:}
We compute a local signal per sample by leveraging label-based similarity between training instances. Given a binary label matrix $Y \in \{0,1\}^{B \times L}$ for a batch of $B$ samples and $L$ labels, we define:

\begin{equation}
\text{KNN}_{\text{raw}} = Y Y^\top
\end{equation}

This yields a $B \times B$ matrix where entry $(i,j)$ counts the number of shared labels between sample $i$ and sample $j$, effectively forming a sample-wise co-occurrence affinity. We normalize each row by the label count of the corresponding sample:

\begin{equation}
\text{KNN}_{\text{norm}}[i, j] = \frac{\text{KNN}_{\text{raw}}[i, j]}{\sum_k Y[i, k] + \varepsilon}
\end{equation}

Finally, we propagate this similarity back into the label space by computing a weighted average over all sample label vectors:

\begin{equation}
\text{KNN}_l = \text{KNN}_{\text{norm}} \cdot Y
\end{equation}

The result $\text{KNN}_l \in \mathbb{R}^{B \times L}$ is a soft, dense score matrix aligned with model output logits, where each entry reflects the relative prevalence of label $l$ among samples similar to the given instance. This method performs a differentiable, soft KNN operation entirely within the label space, without relying on learned embeddings or feature distances.

Where $\lambda_m = 0.1$. The margin loss term is defined as:
\begin{equation}
\text{MarginLoss}(z_l, \theta_l, y_l) =
\begin{cases}
\max(0, \theta_l - z_l + \Delta), & \text{if } y_l = 1 \\
\max(0, z_l - \theta_l + \Delta), & \text{if } y_l = 0
\end{cases}
\end{equation}
We use $\Delta = 0.1$ in all experiments.

\textbf{Intuition:} Margin loss penalizes uncertain predictions near the threshold, improving boundary sharpness. This is especially useful for rare labels or cases where logits tend to hover near the threshold, helping avoid indecisive predictions and improving calibration.

\textbf{Penalization vs. Rewarding:} A central component of our method is the use of thresholds as penalization terms. Instead of encouraging the model to push logits higher (as in reward-based designs), we subtract the threshold from logits before computing the loss. This discourages false positives without disproportionately inflating strong activations. Penalization provides more nuanced control, especially for low-frequency labels that require activation only in highly confident settings.

\textbf{Note on Global Signal:}
The global signal is computed using an IDF-style rarity prior:
\begin{equation}
\text{IDF}_l = \log \left( \frac{N}{f_l + \epsilon} \right)
\end{equation}
Here, $f_l$ is the frequency of label $l$ across the dataset, and $N$ is the total number of samples.
This score reflects how uncommon a label is and biases the threshold higher for rare labels. 
Our current implementation uses only the inverse document frequency (IDF) component, based on global label occurrence. It does not incorporate per-instance term frequencies.

In future work, we plan to extend this to a full TF-IDF formulation by integrating instance-aware label frequency statistics, allowing the threshold to reflect both label rarity and local salience per sample.

\section{Experimental Setup}
\textbf{Datasets:}
\begin{itemize}
  \item \textbf{BibTeX} — 159 labels, 7,395 samples
  \item \textbf{Delicious} — 983 labels, 16,105 samples
  \item \textbf{AmazonCat-13K} — 1.18M samples, 13,330 labels. Derived from product titles and category codes. We use Version 1 TF-IDF features from the AttentionXML repository~\citep{you2019attentionxmllabeltreebasedattentionaware} due to their effective compromise between vocabulary richness and memory footprint. This version also reflects a real-world long-tailed label distribution and is validated in multiple XML studies.
\end{itemize}

\textbf{Hardware:} All experiments were conducted using a single NVIDIA RTX 4090 GPU with 200 GB system RAM.

\textbf{Baselines:} Static threshold, IDF only, KNN only.

\textbf{Metrics:} Macro-F1, Micro-F1, positive ratio.

\textbf{Model Size:} The architecture consists of a shallow multilayer perceptron (MLP) with approximately 2.8 million parameters.

\textbf{Training:} All models were initially scheduled to train for 1500 epochs using BCE loss with optional margin regularization and positive weighting (batch size 128). However, two variants—the IDF-only ablation (no-KNN) and the static threshold baseline—converged early and were stopped at 150 epochs to prevent overfitting. The KNN-only ablation and the full adaptive model completed the full 1500 epochs.

\section{Results}
To evaluate our method, we compare macro-F1 performance across four variants: the full adaptive model (with IDF and KNN fusion), two ablations (IDF only and KNN only), and a static threshold baseline using 0.5.

\textbf{Performance Trajectory:} As shown in Figure 1, the adaptive thresholding model exhibits steady performance improvements, surpassing all baselines by a wide margin.

\textbf{Final Metrics:} Figure 2 and Table 1 illustrate the outcome of each method. Notably, the adaptive model achieves a macro-F1 of 0.1712 with the lowest BCE loss and most conservative positive prediction rate, reinforcing the method’s robustness.
This result outperforms prior published macro-F1 scores on AmazonCat-13K by a significant margin—surpassing AttentionXML and DEPL-style methods by over 6 points—while using a simpler architecture without label trees or pretrained language models.

\textbf{Weight Dynamics:} Figure 3 visualizes the progression of learned coefficients $\alpha$, $\beta$, and $\lambda$. The adaptive model learns to prioritize the more reliable signal depending on label rarity and context.

\subsection{F1 Score Across Epochs}
\includegraphics[width=0.95\linewidth]{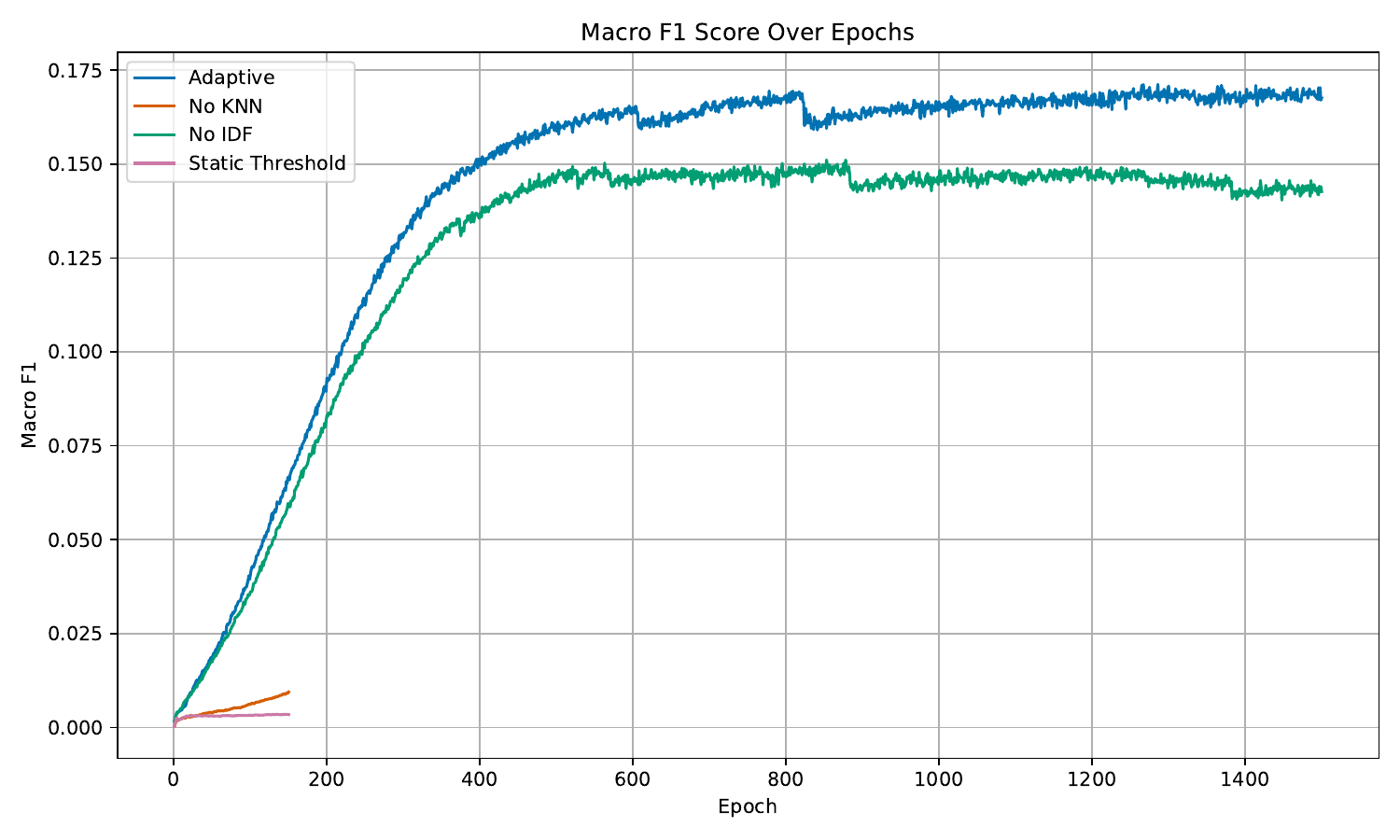}

\textbf{Figure 1:} Macro-F1 score over training epochs for four models. Adaptive and KNN-only variants trained for the full 1500 epochs. IDF-only and static threshold models were stopped at 150 epochs due to early convergence.

\subsection{Final Macro-F1 Comparison}
\includegraphics[width=0.75\linewidth]{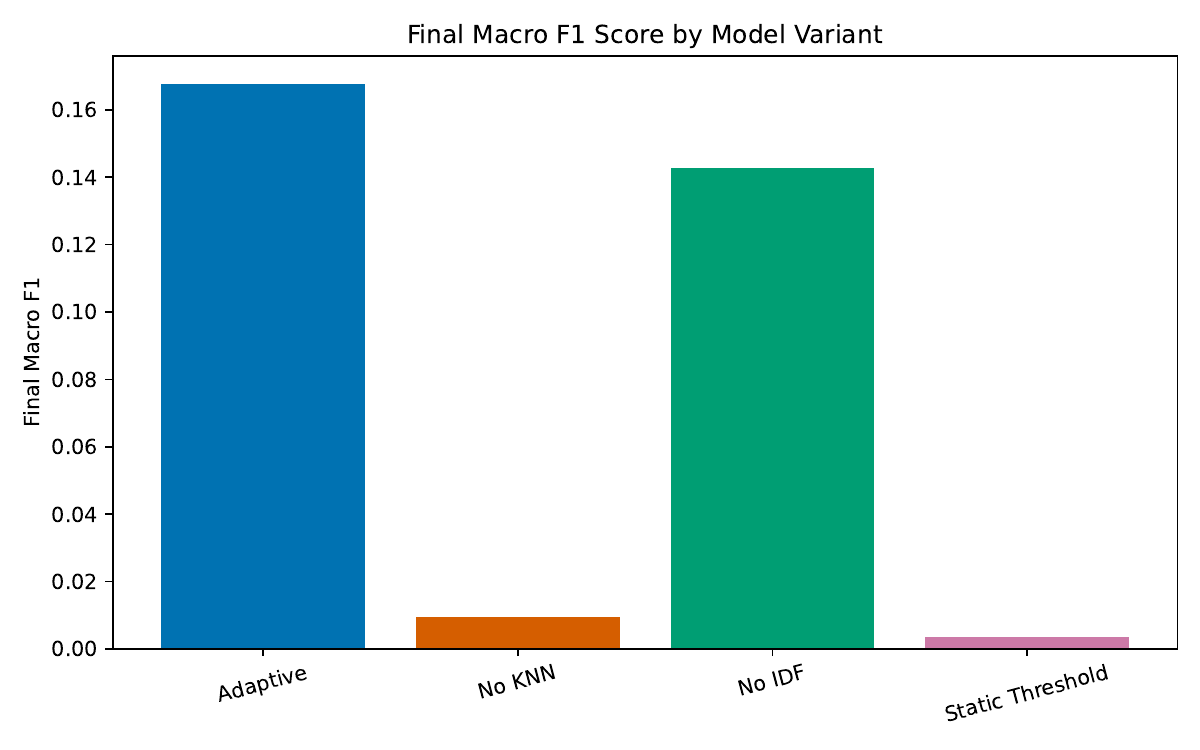}

\textbf{Figure 2:} Final macro-F1 score per model variant. Adaptive model reaches 0.1712. Ablations show each component contributes to performance.

\subsection{Final Metrics Summary}

\begin{table}[htbp]
\centering
\caption{Final metrics on AmazonCat-13K after training. The adaptive model outperforms all baselines across all metrics.}
\label{tab:final_metrics}
\begin{tabular}{lccc}
\toprule
\textbf{Model Variant} & \textbf{Macro F1} & \textbf{BCE Loss} & \textbf{Pos \%} \\
\midrule
Adaptive (IDF+KNN)   & \textbf{0.1712} & 0.3118 & 0.0006 \\
KNN Only (No-IDF ablation)              & 0.1456          & 0.3131 & 0.0007 \\
IDF Only (No-KNN ablation)            & 0.0094          & 0.3197 & 0.0007 \\
Static Threshold (0.5) & 0.0035          & 0.4754 & 0.0023 \\
\bottomrule
\end{tabular}
\end{table}

\subsection{Threshold Weight Evolution}
\includegraphics[width=0.95\linewidth]{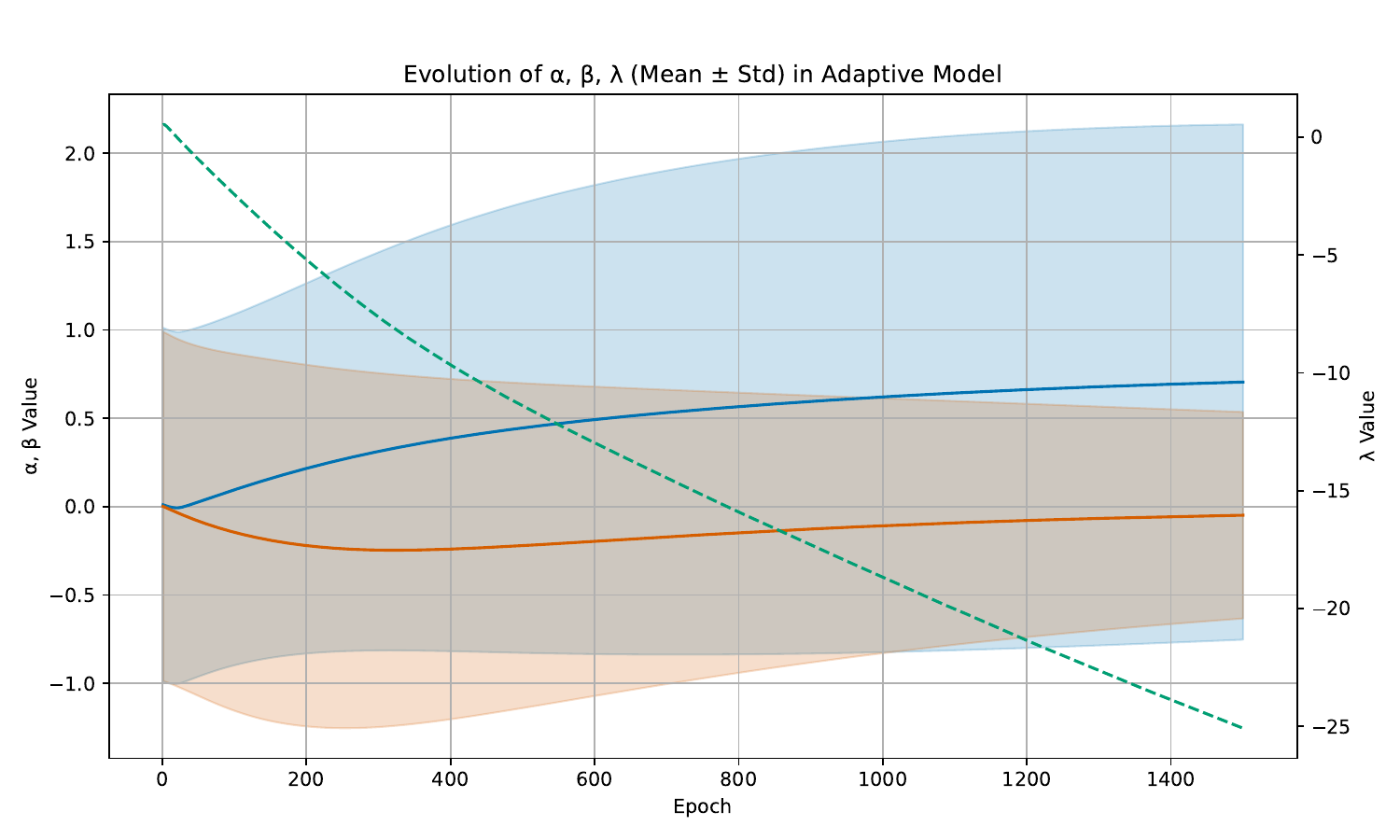}

\textbf{Figure 3:} Mean and standard deviation of learned weights $\alpha$, $\beta$, and blend coefficient $\lambda$ over training. IDF and KNN contributions evolve independently, and $\lambda$ skews toward the stronger signal as learning progresses.

\section{Discussion}
Our adaptive approach demonstrates strong performance across all model variants. The contributions of global (IDF) and local (KNN) signals can be visualized via the learned weights $\alpha$ and $\beta$, while $\lambda$ evolves during training to reflect the model's preference. We observe that $\lambda$ tends to shift toward the KNN signal for rare labels, suggesting increased reliance on local context in low-frequency settings. The margin loss term further sharpens decision boundaries, discouraging uncertain activations and improving calibration.

We hypothesize that adaptive thresholding offers the greatest benefit in datasets with large label spaces and heavy imbalance. In preliminary tests on BibTeX and Delicious, gains were modest, likely due to dense label co-occurrence and lower rarity skew. In contrast, AmazonCat-13K—with over 13,000 labels and long-tailed frequency—highlighted the benefits of our fusion-based approach.

An important observation arises from training dynamics: the IDF-only (No-KNN) ablation and static threshold baseline both converged prematurely, halting at 150 epochs. In contrast, the adaptive model and KNN-only ablation trained to full duration. This suggests that local context via KNN provides a stronger training signal for convergence than global rarity alone. The superior performance of KNN-only over IDF-only further supports this interpretation, highlighting the need for both components in full synergy.

These findings imply that while global rarity is useful for biasing thresholds upward on infrequent labels, local agreement among similar instances may be more essential for optimizing decision boundaries in extreme MLC. It also motivates fallback mechanisms that prioritize IDF in low-neighborhood-density regimes.

\textbf{Comparison to Prior Work:}
Previous studies such as AttentionXML~\citep{you2019attentionxmllabeltreebasedattentionaware} reported macro-F1 scores around 0.07 on AmazonCat-13K.

Other approaches like pseudo-label guided generation~\citep{zhang2022longtailedextrememultilabeltext} improved performance via external semantics, achieving up to 0.11 macro-F1.

Our approach, without leveraging a tree structure or pretrained transformers, achieves 0.1712.
This substantially surpasses existing benchmarks with a lightweight architecture.

\textbf{Interpretability:} TF-IDF/IDF and KNN are both interpretable and explainable signals. This makes our method attractive in domains requiring trust and auditability. Unlike opaque attention weights or deep threshold regressors, our adaptive penalty reflects known label statistics.

\textbf{Modularity and Efficiency:} Our method operates with a lightweight MLP and cached signals. It can be applied as a modular head to existing pretrained models such as BERT or ClinicalBERT without retraining the backbone. This modularity also opens paths for plug-and-play applications in retrieval or summarization.

\textbf{Future Work:} 
In future work, we aim to extend this to a full TF-IDF formulation by integrating instance-aware statistics.

We also plan to explore more advanced alternatives to the KNN signal, such as differentiable clustering or learned neighborhood graphs, to enhance local context modeling. Another promising direction is shifting the thresholding mechanism from label space to the logits space, enabling more direct integration with learned feature distributions.

Finally, we are currently applying this method to clinical datasets such as MIMIC-III, where threshold precision is critical.

We anticipate further gains in macro-F1 as the framework matures, particularly through improved global-local fusion, full TF-IDF modeling, and adaptation to clinical contexts where high precision is paramount.

\section{Conclusion}
We introduce a learnable, interpretable adaptive thresholding layer for MLC. By fusing IDF-based and KNN-derived signals, it learns when to activate each label with respect to global rarity and local structure. This method boosts macro F1 without requiring deep architectures or external pretraining. It opens the door to structured thresholding in medical coding and other safety-critical domains.
This framework is especially well-suited to medical domains where false positives carry serious risk and interpretability is essential. We believe it offers a promising foundation for structured prediction in healthcare NLP, including ICD assignment and clinical summarization.

\section*{Acknowledgments}
The author would like to thank his family for their patience and support throughout the weekends dedicated to developing and validating this work.

\bibliographystyle{plainnat}
\bibliography{references}

\end{document}